\documentclass[conference]{IEEEtran}
\IEEEoverridecommandlockouts
\usepackage{cite}
\usepackage{amsmath,amssymb,amsfonts}
\usepackage{algorithmic}
\usepackage{graphicx}
\usepackage{textcomp}
\usepackage{xcolor}
\usepackage{comment}
\usepackage{adjustbox}
\usepackage{siunitx}
\usepackage{multirow}
\usepackage{booktabs}

\usepackage{xcolor}

\usepackage{threeparttable}
\usepackage{tabu}
\usepackage{caption}
\usepackage{array}
\newcolumntype{$}{>{\global\let\currentrowstyle\relax}}
\newcolumntype{^}{>{\currentrowstyle}}
\newcommand{\rowstyle}[1]{\gdef\currentrowstyle{#1}%
  #1\ignorespaces
}
\definecolor{mycolor}{RGB}{52, 31, 151}

\def\BibTeX{{\rm B\kern-.05em{\sc i\kern-.025em b}\kern-.08em
    T\kern-.1667em\lower.7ex\hbox{E}\kern-.125emX}}
\begin{document}

\title{Sim-To-Real Transfer for Miniature Autonomous Car Racing
}

\author{
\IEEEauthorblockN{
Yeong-Jia Roger Chu \textsuperscript{\rm 1},
Ting-Han Wei \textsuperscript{\rm 4},
Jin-Bo Huang \textsuperscript{\rm 1},
Yuan-Hao Chen \textsuperscript{\rm 1},
I-Chen Wu \textsuperscript{\rm 1,2,3}
}
\IEEEauthorblockA{
\textsuperscript{\rm 1}Department of Computer Science, National Chiao-Tung University, Hsinchu, Taiwan\\
\textsuperscript{\rm 2}Pervasive Artificial Intelligence Research (PAIR) Labs, Taiwan\\
\textsuperscript{\rm 3}Research Center for Information Technology Innovation, Academia Sinica, Taiwan \\
\textsuperscript{\rm 4}Department of Computing Science, University of Alberta, Edmonton, Canada\\
Email: icwu@cs.nctu.edu.tw
}
}

\maketitle

\begin{abstract}
Sim-to-real, a term that describes where a model is trained in a simulator then transferred to the real world, is a technique that enables faster deep reinforcement learning (DRL) training. However, differences between the simulator and the real world often cause the model to perform poorly in the real world. Domain randomization is a way to bridge the sim-to-real gap by exposing the model to a wide range of scenarios so that it can generalize to real-world situations. However, following domain randomization to train an autonomous car racing model with DRL can lead to undesirable outcomes. Namely, a model trained with randomization tends to run slower; a higher completion rate on the testing track comes at the expense of longer lap times. This paper aims to boost the robustness of a trained race car model without compromising racing lap times. For a training track and a testing track having the same shape (and same optimal paths), but with different lighting, background, etc., we first train a model (teacher model) that overfits the training track, moving along a near optimal path. We then use this model to teach a student model the correct actions along with randomization. With our method, a model with 18.4\% completion rate on the testing track is able to help teach a student model with 52\% completion. Moreover, over an average of 50 trials, the student is able to finish a lap 0.23 seconds faster than the teacher. This 0.23 second gap is significant in tight races, with lap times of about 10 to 12 seconds.
\end{abstract}

\begin{IEEEkeywords}
reinforcement learning, deep learning, miniature autonomous race car, sim-to-real, AWS DeepRacer 
\end{IEEEkeywords}

\section{Introduction}

Deep Reinforcement learning (DRL) has shown great success in learning control policies in complex environments without human expert knowledge in many tasks, e.g., object grasping\cite{kalashnikov2018qtopt}, door opening\cite{Gu_2017}, quadrotor flying\cite{PI2020104222}, quadruped robot control\cite{Tan_2018}, etc. 
However, DRL often requires a tremendous amount of experience episodes for learning. Consequently, training DRL for real-world applications tends to be very expensive and time consuming. As an example, Kalashnikov et al. \cite{kalashnikov2018qtopt} collected 580,000 samples over the span of four months using 7 expensive robot arms. In addition to the high time and monetary costs, random exploration is often necessary to obtain optimal or near optimal policies, which can be dangerous and impossible to perform during real-world data collection for tasks like autonomous driving.

\begin{figure}[t]
  \centering
  \includegraphics[width=.8\linewidth]{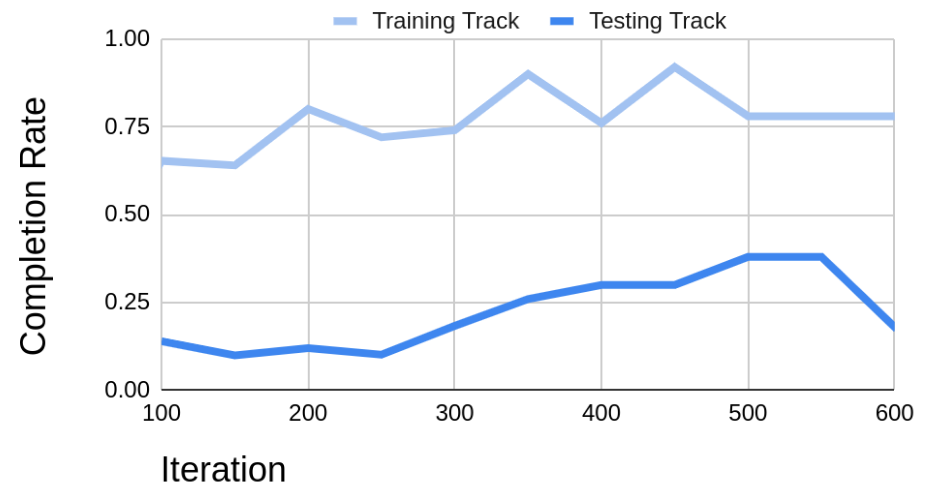}
  \caption{Completion rate of the same model, racing on the training track and the testing track. The horizontal axis represents the iterations of DRL training. This figure shows the performance drop when a model is transferred from the training environment to a different testing environment. At best, the testing track performance still had a 40\% difference, which occurs in iteration 550.}
  \label{figure:sim-to-real-complate-rate}
\end{figure}

Sim-to-real provides a more economic and efficient way of training DRL \cite{Tobin_2017,Peng_2018}, where a replica of the real-world environment is built in a simulator for training. The simulator-trained model is then transferred to the real-world to perform tasks. This enables higher efficiency and scalability for DRL training.

For miniature autonomous car racing, DRL is a suitable way to discover the optimal racing path without human knowledge. AWS DeepRacer \cite{deepracer}, a miniature autonomous car racing platform, also adopts sim-to-real transfer for training DRL models. However, modern physics simulators are still struggling to perfectly replicate the real-world. This reality gap often causes a performance drop when a simulator-trained DRL model is transferred to the real-world. 
Fig. \ref{figure:sim-to-real-complate-rate} shows that there is a large drop in completion rate when the same model is transferred from one track for training to another track for testing, despite them being nearly identical. More specifically, while the shape of the tracks (and also the optimal paths) are the same, various elements of the two environments are different, e.g., lighting, background, track texture, etc, as shown in Fig. \ref{figure:trainig-testing-track}. These elements contributed to completion rate drops throughout the entire training process. At best, the testing track performance still had a 40\% difference, which occurs in iteration 550 shown in Fig. \ref{figure:sim-to-real-complate-rate}.
For more efficient evaluations, we constructed a simulated testing environment with real-world images as the background and different lighting and track textures than the training track. Although the simulated testing track does not fully represent the real world, judging from our results detailed in this paper, we are optimistic about extending these techniques to real-world tests in future investigations.

Domain randomization was first introduced in \cite{Tobin_2017} to bridge the sim-to-real gap in an object localization and grasping task. Various properties of the environment e.g., object texture, lighting, camera position, etc., are randomized so that the trained model is exposed to a wide range of scenarios. By doing so, it can generalize to real-world situations without additional training. Domain randomization was further used in various tasks such as flying a quadcopter \cite{Sadeghi_2017}, autonomous driving \cite{Tremblay_2018}, robot arm control \cite{Peng_2018}, etc.

\begin{figure}[t]
  \centering
  \includegraphics[width=1\linewidth]{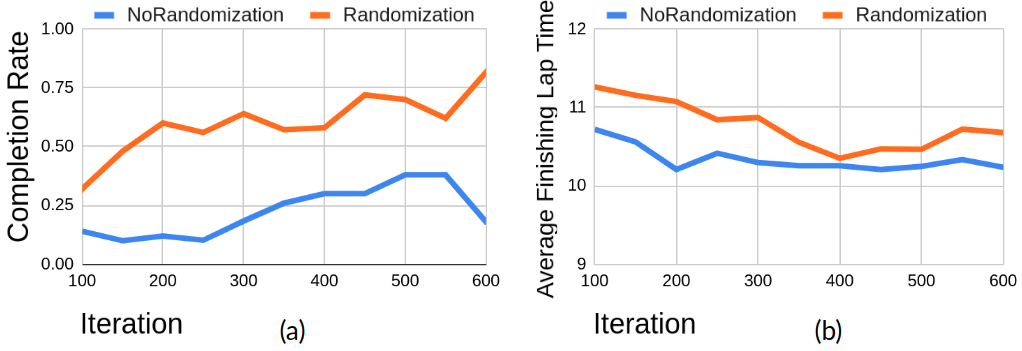}
  \caption{Completion rate and average lap times on the testing track for each DRL training iteration. 
  The model trained with randomization is able to achieve higher completion rates, shown in (a), but it prefers to drive more conservatively and finishes a lap slower, shown in (b). Higher completion rate on the testing track comes at the expense of longer lap times, which runs counter to the central goal of racing.}
  \label{figure:rand-norand-completion-rate-lap-time-testing-track}
\end{figure}

However, applying random transformations on observations, to train an autonomous car racing model with DRL can lead to undesirable outcomes. Namely, with domain randomization, a higher completion rate on the testing track comes at the expense of longer lap times. Fig. \ref{figure:rand-norand-completion-rate-lap-time-testing-track} (a) shows that the models trained with randomization is able to achieve higher completion rate on the testing track. But from Fig. \ref{figure:rand-norand-completion-rate-lap-time-testing-track} (b), we can see that the model prefers to drive more conservatively, which runs counter to the central goal of racing.

This paper aims to boost the robustness of a trained race car model without compromising racing lap times. Instead of training a model with DRL along with randomization, we first train a model without any random transformations resulting in a model that overfits the training track. We refer to this model as the \textit{teacher}, since its policy is near optimal. We then perform behavior cloning along with randomization to train a \textit{student} model that runs as fast as the teacher, and with higher robustness to environment differences.
The method consists of the following three steps.
(1) An original observation is fed into the teacher model to obtain an action,
(2) perform randomization on the original observation, (3) feed the randomized observation into the student model and force it to output the same action as the teacher.
With our method, a teacher model with an 18.4\% completion rate on the testing track can help train a student model that achieves 52\% completion. Moreover, over an average of 50 trials the student is able to finish a lap 0.23 seconds faster than the teacher. With a typical lap time of 10 to 12 seconds, this 0.23 second gap is significant for tight races.

The paper is organized as follows. Section \ref{background} describes the background and lists related works. Section \ref{method} introduces our method. Section \ref{experiments} discusses experiments to evaluate the performance of our method. Finally, Section \ref{conclusion} presents our conclusions.

\section{Background \label{background}}
\subsection{Deep Reinforcement Learning}
Deep reinforcement learning has demonstrated many successes, e.g., AlphaGo\cite{alphago} (for the game of Go), and Deep Q-Network (DQN)\cite{dqn} (for Atari games), among others; DRL has also been successfully applied to various robotic tasks, e.g., object grasping\cite{kalashnikov2018qtopt}, door opening\cite{Gu_2017}, quadrotor flying\cite{PI2020104222}, quadruped robot control\cite{Tan_2018}, etc. Proximal policy optimization (PPO) \cite{ppo} is a well-known DRL algorithm that has successfully learned complicated control tasks \cite{Andrychowicz_2019}.
In this paper, we use PPO to train the teacher models.

\subsection{Miniature Autonomous Car Racing}
Miniature autonomous car racing provides a more affordable way to conduct research compared to real-sized autonomous race cars. F1/10 \cite{F1tenth} is an open-source platform for miniature autonomous car racing. It provides software tools and instructions for hardware setups of a 1/10th scale race car equipped with various sensors, e.g., LiDAR, RGBD cameras.

DeepRacer\cite{deepracer} is an autonomous car racing platform for end-to-end DRL supported by Amazon Web Services (AWS). In the DeepRacer environment, the model's goal is to finish a lap on a race track as fast as possible. The models take as input a single image captured by a front-facing camera on the race car. Each image is converted to a $120\times160$ grayscale image and normalized to [0, 1] before passing it to the model. DeepRacer adopts discrete action spaces with four customizable parameters:  maximum speed, maximum angle, speed granularity and angle granularity. For example, setting \textit{maximum speed} $= 6 m/s$, \textit{maximum angle} $= 20^{\circ}$, \textit{speed granularity} $= 3$ and \textit{angle granularity} $= 5$ results in three speeds \{2 m/s, 4 m/s, 6 m/s\} and 5 angles \{left 20\textdegree, left 10\textdegree, 0\textdegree, right 10\textdegree, right 20\textdegree\}. Each angle is paired with each speed to form a discrete action space of 15 actions. In this paper, we conduct experiments on the DeepRacer platform.
\subsection{Sim-to-real Transfer}
Sim-to-real enables faster DRL training where a model is trained with simulated data and transferred to the real world to perform tasks. However, the differences between the simulator and the real world often cause the model to perform poorly in the real world. In a problem setting where real-world data can be accessed during training, many works have successfully reduced the sim-to-real performance drop by utilizing real-world data. Rusu et al. \cite{rusu2016simtoreal} utilized progressive networks for fast adaptation to real-world data. Chebotar et al. \cite{Chebotar_2019} adjusted simulator parameters to fit the real world environment by aligning trajectories collected in the simulator to the real world. This paper does not consider using real-world data for training, but it can be extended to this problem setting in the future.

In another problem setting where real-world data cannot be accessed during training, many works have dealt with the sim-to-real problem by exposing the model to a wide range of scenarios so that it can generalize to real-world situations without additional training. Domain randomization was first introduced in \cite{Tobin_2017} to bridge the sim-to-real gap in an object localization and grasping task. Various properties of the environment e.g., object texture, lighting, camera position, etc., are randomized. Dynamics randomization \cite{Peng_2018} was used for robot arm control tasks where a set of physics engine dynamics parameters are randomized, e.g., mass of the robot arm, damping, friction, etc. Randomization techniques were further used in various tasks such as flying a quadcopter \cite{Sadeghi_2017}, autonomous driving \cite{Tremblay_2018}, etc. Balaji et al. \cite{deepracer} dealt with the sim-to-real problem specifically for DeepRacer using domain randomization.
The methods mentioned above all trained their policies along with domain randomization, which may come with undesirable side effects for autonomous car racing. Namely, training an autonomous car racing model with domain randomization can result in slower performance, shown in Fig. \ref{figure:rand-norand-completion-rate-lap-time-testing-track}. This runs counter to the central goal of racing. This paper aims to boost the robustness of a trained race car model without compromising racing lap times.

\section{Method\label{method}}
Given a set of observations $O$ collected from the training environment, and a teacher model $\pi_T$ trained on the same environment, our goal is to train a student model $\pi_S$ that can perform as well as $\pi_T$ on the training environment, while simultaneously performing more robustly than $\pi_T$ when transferred to other environments. In this paper, each observation is a single image capture by the front-facing camera of the race car. The models take as input an observation, and output a vector of length $n$. We then pass the output vector through a softmax function to obtain the probability of choosing the $n$ actions.
\subsection {Design and Loss Function}
\begin{figure}[ht]
  \centering
  \includegraphics[width=.8\linewidth]{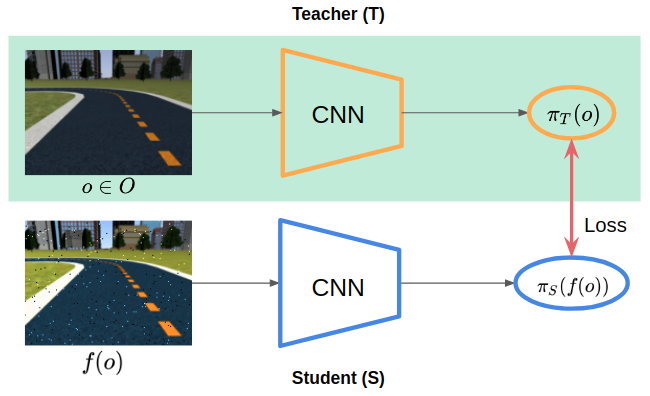}
  \caption{The design of our method.}
  \label{figure:design}
\end{figure}
The design of our method is shown in Fig. \ref{figure:design}. First, an observation $o$ is randomly sampled from $O$, and fed into the teacher model to obtain the teacher output $\pi_T(o)$. Then, a set of predefined randomization functions $f$ are performed on $o$, resulting in a randomized observation $f(o)$. Finally, we force the student to output the same action as the teacher by training the student model with Huber-style loss similar to \cite{Sermanet_2018}:
\begin{equation}
Loss = \Vert \pi_T(o) - \pi_S(f(o))\Vert_2^2 + \Vert \pi_T(o) - \pi_S(f(o)) \Vert_2
\end{equation}
$\pi_T(o)$ and $\pi_S(f(o))$ are the output vectors (for the $n$ actions) before the softmax layer.
The squared Euclidean distance (the first term) provides steeper gradients for early stages of the training when $\pi_T(o)$ and $\pi_S(f(o))$ are further apart. The second term starts prevailing when $\pi_T(o)$ and $\pi_S(f(o))$ are very close and provides finer gradients at later stages of training.

\subsection {Randomization Functions}
\begin{figure}[t]
  \centering
  \includegraphics[width=.88\linewidth]{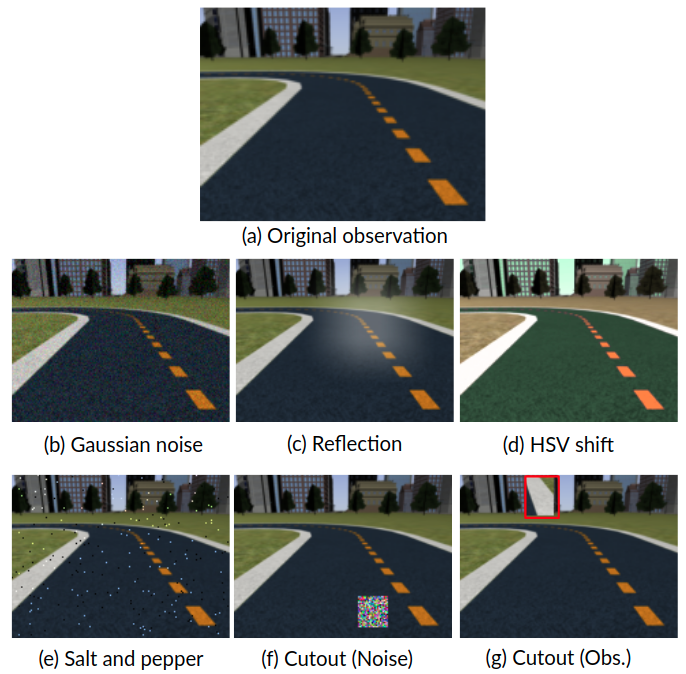}
  \caption{Various randomization effects added to observations, where (a) is the original observation. For (g), the red box is not part of the randomized image, and is only added to indicate the cutout area.}
  \label{figure:randomization-methods}
\end{figure}

We perform randomization on observations of size $120\times160$ with three channels, and the pixel value ranges from 0 to 255.
Our method is based on the assumption that the actions provided by the teacher for the original observations still apply to the randomized ones. For this assumption to hold, the randomization functions should be carefully chosen. 
In this paper, we define six randomization functions, shown in Fig. \ref{figure:randomization-methods}, where all six are different ways of adding noise to the observation.
\begin{enumerate}
    \item Gaussian noise \cite{pmlr-v15-bengio11b}: for each pixel, add a noise that is sampled from the Gaussian distribution with $\mu = 0, \sigma = 10$.
    \item Reflection: add a simulated oval-shaped reflection to a random position.
    \item HSV shift: multiply each channel in the input's HSV space with a random value uniformly sampled from range [0.5, 1.5]. The values exceeding the channel value limits are clipped.  
    \item Salt and pepper noise \cite{pmlr-v15-bengio11b}: set the lightness of randomly selected pixels to the highest or lowest value.
    \item Cutout (Noise) \cite{devries2017improved}: randomly remove a rectangular area of the observation and fill it with random noise.
    \item Cutout (Obs.): randomly remove a rectangular area and fill it with random areas taken from training track observations $O$. The cutout area is always located on the upper part of the observation, so that the model can learn to ignore the background.
\end{enumerate}
None of the randomization functions include shift, rotation or translation, because these functions are prone to changing the optimal action.

\section {Experiments\label{experiments}}
\subsection {Experimental Setup}
\subsubsection{Environment}
\begin{figure}[t]
  \centering
  \includegraphics[width=.88\linewidth]{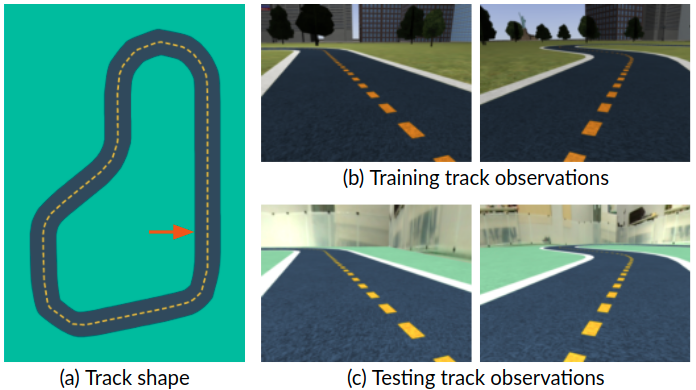}
  \caption{(a) The track shape for both the training and testing tracks. The starting position is indicated by the red arrow. (b) Observations captured on the training track. (c) Observations captured on the testing track.}
  \label{figure:trainig-testing-track}
\end{figure}
Our goal is to first train a teacher that learns the near optimal path of a training track, then train a student model that runs as fast as the teacher with higher robustness when transferred to the same track under a different environment. In this paper, we define one training track and one testing track, both of them have the same shape, shown in Fig. \ref{figure:trainig-testing-track} (a), obtained from the AWS DeepRacer github repository \cite{deepracer_env}. 
The training track, shown in Fig. \ref{figure:trainig-testing-track}(b), is placed inside a simulated city with landscapes, buildings, roads, etc. The testing track, shown in Fig. \ref{figure:trainig-testing-track}(c), has the same shape (and also the same optimal paths) as the training track, but various elements differ. More specifically, the lighting and textures are different, and the background is replaced by real-world images taken from a real-world testing environment.
Although we aim to deal with the sim-to-real problem, this paper uses a simulated testing track to enable more efficient evaluation for our methods. Although we cannot guarantee the best performing model on the testing track will perform well in the real world, we show that our method can boost robustness for this simple case. We will then follow the same method and experiments for real-world evaluations in the future.

The observation is taken from the front-facing camera on the race car with a frame rate of 15 FPS.
For the action space, we set \textit{maximum speed} $= 8 m/s$, \textit{maximum angle} $= 30^{\circ}$, \textit{speed granularity} $= 3$ and \textit{angle granularity} $= 7$. Namely, there are seven angles [-30\textdegree, -20\textdegree, -10\textdegree, 0\textdegree, 10\textdegree, 20\textdegree, 30\textdegree] and three speeds [3 m/s, 5 m/s, 8 m/s], and paring angles and speeds forms 21 actions.

\subsubsection{Network Architecture and Model Training}
The teacher and student models share the same network architecture proposed in \cite{humanlevel}, which consists of 3 convolutional layers followed by 2 fully-connected layers. The kernel initialization function is Glorot Uniform for both convolutional layers and the fully-connected layers and the bias terms are initialized to $0$.

A model that overfits the training track is trained using PPO \cite{ppo} for 600 iterations, where the weights of each iteration is checkpointed and further used as teacher models in the following experiments. Detailed training settings are included in the appendix. During the training process of this overfitting model, observations are collected and later used for student training. Around $10^7$ observations were collected but only $10^5$ of them are randomly selected for training the students.
\subsubsection{Evaluation}
The environment is stochastic because within the simulation, while input observations are sent to the model at 15 FPS, there exists small timing errors so that the observation intervals are not exactly on time.
Each model is evaluated on both the training and the testing track for 50 trials. The race car is placed at the same starting position for each trial, shown in Fig. \ref{figure:trainig-testing-track} (a), indicated by the red arrow.
Models are evaluated with two metrics: completion rate and average lap times. 

For a model to be better than another, it has to achieve a higher completion rate and lower average lap time simultaneously. If a model performs better in one metric but worse in the other, e.g., higher completion rate (better) and longer lap times (worse), there is no clear winner.

\subsection{Experimental Results}
\begin{center}
\begin{table}[ht]
\setlength\extrarowheight{1pt}
\resizebox{\columnwidth}{!}{
\begin{tabular}{ $c ^c ^c ^c c}
 \toprule
 Teacher    & \multirow{2}{*}{Models}  & Completion & Avg. lap    & Min. lap \\
 checkpoint &                          & rate (\%)       & time (sec.) & time (sec.) \\
 
 \midrule
 \multirow{3}{*}{300} & Teacher & 18.4 $\pm$ 10.8 & 10.30 $\pm$ 0.05 & 10.03 \\
                      & S (baseline) & 18.0 $\pm$ 10.6 & 10.12 $\pm$ 0.05 & 9.77 \\
                      \rowstyle{\bfseries} & Student  & 38.8 $\pm$ 13.6 & 10.22 $\pm$ 0.09 & 9.50 \\
 \midrule
 \multirow{3}{*}{450} & Teacher & 30.0 $\pm$ 12.7 & 10.21 $\pm$ 0.10 & 9.50 \\
                      & S (baseline) & 26.0 $\pm$ 12.2 & 10.14 $\pm$ 0.09 & 9.77 \\
                      \rowstyle{\bfseries} & Student  & 42.0 $\pm$ 13.7 & 10.19 $\pm$ 0.08 & 9.44 \\
 \midrule
 \multirow{3}{*}{600} & Teacher & 18.0 $\pm$ 10.6 & 10.24 $\pm$ 0.09 & 9.70 \\
                      & S (baseline) & 16.0 $\pm$ 10.2 & 10.10 $\pm$ 0.08 & 9.64 \\
                      \rowstyle{\bfseries} & Student & 36.0 $\pm$ 13.3 & 10.08 $\pm$ 0.10 & 9.64 \\
 \bottomrule
\end{tabular}
}
\caption{The performance for three groups of teacher and student models. S (Baseline) refers to student models trained without any randomization. The better student model is marked in bold.}
\label{table:three-teacher}
\end{table}
\end{center}
For the first experiment, we show that our method works by experimenting on three teacher models. These three models are three checkpoints during the training process of the overfitting model for the training track, at checkpoints 300, 450, and 600. For each teacher model, one student is trained with all of the six randomization functions, and the other student is trained without any randomization to act as a baseline.
Table \ref{table:three-teacher} shows that for all three teacher models, students trained with randomization perform better than their teachers, whereas the baseline students fail to achieve completion rates higher than their teachers.
The student that has the highest completion rate (42\%) is taught by the teacher at checkpoint 450, but it has the least completion rate difference ($+$12\%) compared to its teacher. The other two students with randomization taught by the teacher at checkpoints 300 and 600 have larger completion rate differences of 20\% and 18\% respectively. We conjecture that there is a limit to completion rate one can reach by using randomization methods, therefore, the students trained by a teacher with higher completion rate is likely to show smaller differences.

In Table \ref{table:aug}, we perform ablation evaluations on the six randomization functions. For each teacher model, one student is trained with all six randomization functions (AllRand) while others are trained by removing one of the randomization functions. Most students perform better in terms of both completion rate and average lap times than their teachers. The student without Gaussian noise taught by the teacher at checkpoint 300 achieves the highest completion rate (52\%) and the most completion difference (34\%) than its teacher. Furthermore, it has an average lap time of 10.07 seconds and a minimum lap time of 9.37 seconds, which is 0.23 seconds and 0.66 seconds faster than its teacher. These lap time gaps are significant for tight races.

\begin{center}
\begin{table}[ht]
\setlength\extrarowheight{1pt}
\centering
\resizebox{\columnwidth}{!}{
\begin{tabular}{ $c ^l ^c ^c c} 
 \toprule
Teacher  & \multirow{2}{*}{Models}  & Completion & Avg. lap    & Min. lap \\
checkpoint &   & rate (\%)       & time (sec.) & time (sec.) \\
 \midrule
 \multirow{8}{*}{300} & Teacher & 18.4 $\pm$ 10.8 & 10.30 $\pm$ 0.05 & 10.03 \\
                      & AllRand & 38.8 $\pm$ 13.6 & 10.22 $\pm$ 0.09 & 9.50 \\
                      \cmidrule(lr){2-5}
                     \rowstyle{\bfseries} & w/o Gaussian & 52.0 $\pm$ 13.8 & 10.07 $\pm$ 0.09 & 9.37 \\
                     \rowstyle{\bfseries} & w/o Reflection & 46.0 $\pm$ 13.8 & 10.02 $\pm$ 0.09 & 9.57 \\
                     \rowstyle{\bfseries} & w/o HSV & 42.9 $\pm$ 13.9 & 10.09 $\pm$ 0.09 & 9.44 \\
                     & w/o SaltPepper & 46.9 $\pm$ 14.0 & 10.23 $\pm$ 0.10 & 9.64 \\
                     & w/o Cutout (Noise) & 36.0 $\pm$ 13.3 & 10.04 $\pm$ 0.08 & 9.44 \\
                      \rowstyle{\bfseries \color{mycolor}} & w/o Cutout (Obs.) & 25.0 $\pm$ 12.2 & 10.25 $\pm$ 0.10 & 9.64 \\
 
 \midrule
  \multirow{8}{*}{450} & Teacher & 30.0 $\pm$ 12.7 & 10.21 $\pm$ 0.10 & 9.50 \\
                     & AllRand & 42.0 $\pm$ 13.7 & 10.19 $\pm$ 0.08 & 9.44 \\
                     \cmidrule(lr){2-5}
                     \rowstyle{\bfseries} & w/o Gaussian & 44.0 $\pm$ 13.8 & 10.18 $\pm$ 0.11 & 9.57 \\
                     \rowstyle{\bfseries} & w/o Reflection & 42.9 $\pm$ 13.9 & 10.16 $\pm$ 0.07 & 9.70 \\
                     \rowstyle{\bfseries} & w/o HSV & 44.0 $\pm$ 13.8 & 10.17 $\pm$ 0.08 & 9.70 \\
                     & w/o SaltPepper & 36.7 $\pm$ 13.5 & 10.04 $\pm$ 0.09 & 9.57 \\
                     & w/o Cutout (Noise) & 34.7 $\pm$ 13.3 & 10.02 $\pm$ 0.09 & 9.57 \\
                     \rowstyle{\bfseries \color{mycolor}} & w/o Cutout (Obs.) & 26.5 $\pm$ 12.4 & 10.32 $\pm$ 0.09 & 9.57 \\
\midrule 
 \multirow{8}{*}{600} & Teacher & 18.0 $\pm$ 10.6 & 10.23 $\pm$ 0.09 & 9.70 \\
                     & AllRand & 36.0 $\pm$ 13.3 & 10.08 $\pm$ 0.10 & 9.64 \\
                     \cmidrule(lr){2-5}
                     & w/o Gaussian & 28.0 $\pm$ 12.4 & 10.04 $\pm$ 0.06 & 9.64 \\
                     & w/o Reflection & 36.7 $\pm$ 13.5 & 10.09 $\pm$ 0.06 & 9.70 \\
                     & w/o HSV & 42.0 $\pm$ 13.7 & 10.17 $\pm$ 0.12 & 9.44 \\
                      \rowstyle{\bfseries \color{mycolor}} & w/o SaltPepper & 26.0 $\pm$ 12.2 & 10.14 $\pm$ 0.06 & 9.90 \\
                      \rowstyle{\bfseries \color{mycolor}} & w/o Cutout (Noise) & 24.0 $\pm$ 11.8 & 10.20 $\pm$ 0.09 & 9.64 \\
                      \rowstyle{\bfseries \color{mycolor}} & w/o Cutout (Obs.) & 20.4 $\pm$ 11.3 & 10.19 $\pm$ 0.06 & 9.77 \\

 \bottomrule
\end{tabular}
}
\caption{Ablation study for randomization functions. For each teacher model, students performing better than AllRand (the student trained with all randomization functions) in terms of both completion rate and average lap times are marked in bold. The students performing worse than AllRand are marked in purple. Other students are not directly labeled since they perform better in one metric but worse in the other.}
\label{table:aug}
\end{table}
\end{center}

The student without cutout (Obs.) taught by the teacher at checkpoint 450 is the only student that performs worse than its teacher. Moreover, for all three teacher models, the students without cutout (Obs.) performs worse than the AllRand models. This indicates that cutout (Obs.) is crucial for domain transfer to the testing track. For two out of three teacher models (checkpoint 300 and 450), students trained without Gaussian noise, reflection and HSV shift perform better than AllRand. We conjecture that applying these three randomizations results in a distribution of observations that are far away from observations of the testing track, leading to negative effects. To definitively come to a conclusion regarding the different randomizations, we would need to perform additional experiments by altering properties for the testing track (e.g., texture, lighting, background, etc.). Under different testing track environments, a different combination of randomization functions may be more suitable. Nonetheless, we can follow the same training process and experiments with different randomization functions to find the best student model for other testing tracks, e.g., real-world tracks.

\begin{figure}[t]
  \centering
  \resizebox{\columnwidth}{!}{
  \includegraphics[width=1\linewidth]{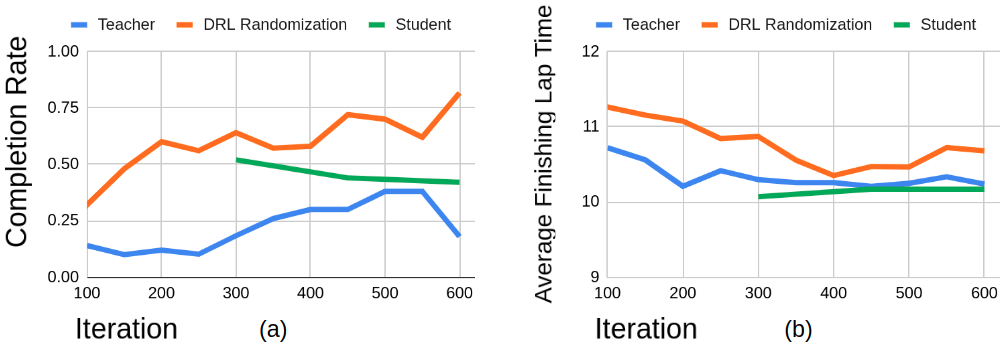}
  }
  \caption{Completion rate and average finish lap times on the simulated testing track.
   The student curve consists of three data points from Table \ref{table:aug}: w/o Gaussian at checkpoint 300, w/o HSV at checkpoint 450, and w/o HSV at checkpoint 600.}
  \label{figure:sim-final}
\end{figure}

We compare the three methods, (1) DRL trained along with randomization, (2) DRL trained without randomization (teacher), and (3) our method, on the testing track in Fig. \ref{figure:sim-final}. The result shows that our method is able to obtain models with the lowest average lap times. Although the completion rate of the student model does not exceed models trained with DRL along with randomization, our method is able to obtain improved versions of the teacher model. This is desirable in tight races where slower lap times are unable to win.

\section{Conclusion\label{conclusion}}
 We demonstrate that our method is able to boost robustness of DRL autonomous car racing models without compromising racing lap times. Namely, a model overfitting the training track can be used to teach a student model that performs better in terms completion rate and racing lap times on the testing track. In one case, one of the teacher models with an 18.4\% completion rate on the testing track can help train a student model that achieves 52\% completion. Moreover, the student is able to finish a lap 0.23 seconds faster than the teacher over an average of 50 trials, which can be a significant gap for tight races.
 Although a simulated testing track is used in this paper for efficient evaluation, we can follow the same method for real-world evaluation in the future.

 This paper shows that our method works for models that take as input a single observation. For models that takes as input multiple observations or recurrent neural networks, our method should work with some modifications. More specifically, observations for training student models should be collected using its teacher models, and observations should be saved in a way that we can retrieve the full trajectories.

 In this paper, we assume there is no access to the testing track observations for training. However, in a problem setting where we can collect data on the testing track, we can adopt generative adversarial nets \cite{goodfellow2014generative} to adapt to the testing track. More specifically, we can replace randomization functions with CycleGAN \cite{Zhu_2017} to translate observations from the training track to testing track image styles.

\section*{Acknowledgment}
This research is partially supported by the Ministry of Science and Technology (MOST) of Taiwan under Grant Number 109-2634-F-009-019 through Pervasive Artificial Intelligence Research (PAIR) Labs.  The computing resource is partially supported by National Center for High-performance Computing (NCHC) of Taiwan.

\bibliographystyle{IEEEtran}
\bibliography{reference}

\begin{thebibliography}{10}
\providecommand{\url}[1]{#1}
\csname url@samestyle\endcsname
\providecommand{\newblock}{\relax}
\providecommand{\bibinfo}[2]{#2}
\providecommand{\BIBentrySTDinterwordspacing}{\spaceskip=0pt\relax}
\providecommand{\BIBentryALTinterwordstretchfactor}{4}
\providecommand{\BIBentryALTinterwordspacing}{\spaceskip=\fontdimen2\font plus
\BIBentryALTinterwordstretchfactor\fontdimen3\font minus
  \fontdimen4\font\relax}
\providecommand{\BIBforeignlanguage}[2]{{%
\expandafter\ifx\csname l@#1\endcsname\relax
\typeout{** WARNING: IEEEtran.bst: No hyphenation pattern has been}%
\typeout{** loaded for the language `#1'. Using the pattern for}%
\typeout{** the default language instead.}%
\else
\language=\csname l@#1\endcsname
\fi
#2}}
\providecommand{\BIBdecl}{\relax}
\BIBdecl

\bibitem{kalashnikov2018qtopt}
D.~Kalashnikov, A.~Irpan, P.~Pastor, J.~Ibarz, A.~Herzog, E.~Jang, D.~Quillen,
  E.~Holly, M.~Kalakrishnan, V.~Vanhoucke, and S.~Levine, ``Qt-opt: Scalable
  deep reinforcement learning for vision-based robotic manipulation,'' 2018.

\bibitem{Gu_2017}
\BIBentryALTinterwordspacing
S.~Gu, E.~Holly, T.~Lillicrap, and S.~Levine, ``Deep reinforcement learning for
  robotic manipulation with asynchronous off-policy updates,'' \emph{2017 IEEE
  International Conference on Robotics and Automation (ICRA)}, May 2017.
  [Online]. Available: \url{http://dx.doi.org/10.1109/ICRA.2017.7989385}
\BIBentrySTDinterwordspacing

\bibitem{PI2020104222}
\BIBentryALTinterwordspacing
C.-H. Pi, K.-C. Hu, S.~Cheng, and I.-C. Wu, ``Low-level autonomous control and
  tracking of quadrotor using reinforcement learning,'' \emph{Control
  Engineering Practice}, vol.~95, p. 104222, 2020. [Online]. Available:
  \url{http://www.sciencedirect.com/science/article/pii/S0967066119301923}
\BIBentrySTDinterwordspacing

\bibitem{Tan_2018}
\BIBentryALTinterwordspacing
J.~Tan, T.~Zhang, E.~Coumans, A.~Iscen, Y.~Bai, D.~Hafner, S.~Bohez, and
  V.~Vanhoucke, ``Sim-to-real: Learning agile locomotion for quadruped
  robots,'' \emph{Robotics: Science and Systems XIV}, Jun 2018. [Online].
  Available: \url{http://dx.doi.org/10.15607/RSS.2018.XIV.010}
\BIBentrySTDinterwordspacing

\bibitem{Tobin_2017}
\BIBentryALTinterwordspacing
J.~Tobin, R.~Fong, A.~Ray, J.~Schneider, W.~Zaremba, and P.~Abbeel, ``Domain
  randomization for transferring deep neural networks from simulation to the
  real world,'' \emph{2017 IEEE/RSJ International Conference on Intelligent
  Robots and Systems (IROS)}, Sep 2017. [Online]. Available:
  \url{http://dx.doi.org/10.1109/IROS.2017.8202133}
\BIBentrySTDinterwordspacing

\bibitem{Peng_2018}
\BIBentryALTinterwordspacing
X.~B. Peng, M.~Andrychowicz, W.~Zaremba, and P.~Abbeel, ``Sim-to-real transfer
  of robotic control with dynamics randomization,'' \emph{2018 IEEE
  International Conference on Robotics and Automation (ICRA)}, May 2018.
  [Online]. Available: \url{http://dx.doi.org/10.1109/ICRA.2018.8460528}
\BIBentrySTDinterwordspacing

\bibitem{deepracer}
B.~Balaji, S.~Mallya, S.~Genc, S.~Gupta, L.~Dirac, V.~Khare, G.~Roy, T.~Sun,
  Y.~Tao, B.~Townsend, E.~Calleja, S.~Muralidhara, and D.~Karuppasamy,
  ``Deepracer: Educational autonomous racing platform for experimentation with
  sim2real reinforcement learning,'' 2019.

\bibitem{Sadeghi_2017}
\BIBentryALTinterwordspacing
F.~Sadeghi and S.~Levine, ``Cad2rl: Real single-image flight without a single
  real image,'' \emph{Robotics: Science and Systems XIII}, Jul 2017. [Online].
  Available: \url{http://dx.doi.org/10.15607/RSS.2017.XIII.034}
\BIBentrySTDinterwordspacing

\bibitem{Tremblay_2018}
\BIBentryALTinterwordspacing
J.~Tremblay, A.~Prakash, D.~Acuna, M.~Brophy, V.~Jampani, C.~Anil, T.~To,
  E.~Cameracci, S.~Boochoon, and S.~Birchfield, ``Training deep networks with
  synthetic data: Bridging the reality gap by domain randomization,''
  \emph{2018 IEEE/CVF Conference on Computer Vision and Pattern Recognition
  Workshops (CVPRW)}, Jun 2018. [Online]. Available:
  \url{http://dx.doi.org/10.1109/CVPRW.2018.00143}
\BIBentrySTDinterwordspacing

\bibitem{alphago}
D.~Silver, A.~Huang, C.~Maddison, A.~Guez, L.~Sifre, G.~Driessche,
  J.~Schrittwieser, I.~Antonoglou, V.~Panneershelvam, M.~Lanctot, S.~Dieleman,
  D.~Grewe, J.~Nham, N.~Kalchbrenner, I.~Sutskever, T.~Lillicrap, M.~Leach,
  K.~Kavukcuoglu, T.~Graepel, and D.~Hassabis, ``Mastering the game of go with
  deep neural networks and tree search,'' \emph{Nature}, vol. 529, pp.
  484--489, 01 2016.

\bibitem{dqn}
V.~Mnih, K.~Kavukcuoglu, D.~Silver, A.~Graves, I.~Antonoglou, D.~Wierstra, and
  M.~Riedmiller, ``Playing atari with deep reinforcement learning,'' 2013.

\bibitem{ppo}
J.~Schulman, F.~Wolski, P.~Dhariwal, A.~Radford, and O.~Klimov, ``Proximal
  policy optimization algorithms,'' 2017.

\bibitem{Andrychowicz_2019}
\BIBentryALTinterwordspacing
O.~M. Andrychowicz, B.~Baker, M.~Chociej, R.~Józefowicz, B.~McGrew,
  J.~Pachocki, A.~Petron, M.~Plappert, G.~Powell, A.~Ray, and et~al.,
  ``Learning dexterous in-hand manipulation,'' \emph{The International Journal
  of Robotics Research}, vol.~39, no.~1, p. 3–20, Nov 2019. [Online].
  Available: \url{http://dx.doi.org/10.1177/0278364919887447}
\BIBentrySTDinterwordspacing

\bibitem{F1tenth}
M.~O'Kelly, V.~Sukhil, H.~Abbas, J.~Harkins, C.~Kao, Y.~V. Pant, R.~Mangharam,
  D.~Agarwal, M.~Behl, P.~Burgio, and M.~Bertogna, ``F1/10: An open-source
  autonomous cyber-physical platform,'' 2019.

\bibitem{rusu2016simtoreal}
A.~A. Rusu, M.~Vecerik, T.~Rothörl, N.~Heess, R.~Pascanu, and R.~Hadsell,
  ``Sim-to-real robot learning from pixels with progressive nets,'' 2016.

\bibitem{Chebotar_2019}
\BIBentryALTinterwordspacing
Y.~Chebotar, A.~Handa, V.~Makoviychuk, M.~Macklin, J.~Issac, N.~Ratliff, and
  D.~Fox, ``Closing the sim-to-real loop: Adapting simulation randomization
  with real world experience,'' \emph{2019 International Conference on Robotics
  and Automation (ICRA)}, May 2019. [Online]. Available:
  \url{http://dx.doi.org/10.1109/ICRA.2019.8793789}
\BIBentrySTDinterwordspacing

\bibitem{Sermanet_2018}
\BIBentryALTinterwordspacing
P.~Sermanet, C.~Lynch, Y.~Chebotar, J.~Hsu, E.~Jang, S.~Schaal, S.~Levine, and
  G.~Brain, ``Time-contrastive networks: Self-supervised learning from video,''
  \emph{2018 IEEE International Conference on Robotics and Automation (ICRA)},
  May 2018. [Online]. Available:
  \url{http://dx.doi.org/10.1109/ICRA.2018.8462891}
\BIBentrySTDinterwordspacing

\bibitem{pmlr-v15-bengio11b}
\BIBentryALTinterwordspacing
Y.~Bengio, F.~Bastien, A.~Bergeron, N.~Boulanger–Lewandowski, T.~Breuel,
  Y.~Chherawala, M.~Cisse, M.~Côté, D.~Erhan, J.~Eustache, X.~Glorot,
  X.~Muller, S.~P. Lebeuf, R.~Pascanu, S.~Rifai, F.~Savard, and G.~Sicard,
  ``Deep learners benefit more from out-of-distribution examples,'' in
  \emph{Proceedings of the Fourteenth International Conference on Artificial
  Intelligence and Statistics}, ser. Proceedings of Machine Learning Research,
  G.~Gordon, D.~Dunson, and M.~Dudík, Eds., vol.~15.\hskip 1em plus 0.5em
  minus 0.4em\relax Fort Lauderdale, FL, USA: PMLR, 11--13 Apr 2011, pp.
  164--172. [Online]. Available:
  \url{http://proceedings.mlr.press/v15/bengio11b.html}
\BIBentrySTDinterwordspacing

\bibitem{devries2017improved}
T.~DeVries and G.~W. Taylor, ``Improved regularization of convolutional neural
  networks with cutout,'' 2017.

\bibitem{deepracer_env}
\BIBentryALTinterwordspacing
{AWS Robotics}, ``Aws robomaker sample application deepracer,'' [Accessed: 1-
  Dec- 2019]. [Online]. Available:
  \url{https://github.com/aws-robotics/aws-robomaker-sample-application-deepracer}
\BIBentrySTDinterwordspacing

\bibitem{humanlevel}
\BIBentryALTinterwordspacing
V.~Mnih, K.~Kavukcuoglu, D.~Silver, A.~A. Rusu, J.~Veness, M.~G. Bellemare,
  A.~Graves, M.~Riedmiller, A.~K. Fidjeland, G.~Ostrovski, S.~Petersen,
  C.~Beattie, A.~Sadik, I.~Antonoglou, H.~King, D.~Kumaran, D.~Wierstra,
  S.~Legg, and D.~Hassabis, ``Human-level control through deep reinforcement
  learning,'' \emph{Nature}, vol. 518, no. 7540, pp. 529--533, Feb. 2015.
  [Online]. Available: \url{http://dx.doi.org/10.1038/nature14236}
\BIBentrySTDinterwordspacing

\bibitem{goodfellow2014generative}
I.~J. Goodfellow, J.~Pouget-Abadie, M.~Mirza, B.~Xu, D.~Warde-Farley, S.~Ozair,
  A.~Courville, and Y.~Bengio, ``Generative adversarial networks,'' 2014.

\bibitem{Zhu_2017}
\BIBentryALTinterwordspacing
J.-Y. Zhu, T.~Park, P.~Isola, and A.~A. Efros, ``Unpaired image-to-image
  translation using cycle-consistent adversarial networks,'' \emph{2017 IEEE
  International Conference on Computer Vision (ICCV)}, Oct 2017. [Online].
  Available: \url{http://dx.doi.org/10.1109/ICCV.2017.244}
\BIBentrySTDinterwordspacing

\end{thebibliography}
\vspace{12pt}

\newpage
\appendix
\begin{center}
\begin{table}[ht]
\setlength\extrarowheight{0pt}
\centering
\small
\begin{tabular}{l p{5cm}}
\toprule
DRL algorithm & Proximal Polixy Optimization \cite{ppo} (PPO) \\
Optimizer &  Adam Optimizer\\
Learning rate & Starting from \(0.0005\) and decays every 4M steps until it reaches 12M steps. \\
Gradient clipping norm & 5.0 \\
Discount factor ($\gamma$) &  0.98 \\ 
PPO clipping parameter & 0.15 \\ 
Minibatch size & 2048 \\ 
Number of sgd epoch & 5 \\ 
Entropy coefficient & Linearly decay from 0.01 to 0 in 8M steps \\
Total training iteration & 600 \\
\bottomrule
\end{tabular}
\caption{Teacher training details and hyperparameters.}
\label{table:teacher-training}
\end{table}
\end{center}

\begin{center}
\begin{table}[ht]
\setlength\extrarowheight{0pt}
\centering
\small
\begin{tabular}{l p{5cm}}
\toprule
Optimizer &  Momentum Optimizer \\
Learning rate & Starting from \(0.001\) and decays every 10 iterations \\
Gradient clipping norm & 5.0 \\
Training data size & 100,000 \\
Minibatch size & 500 \\
Total training iteration & 50 \\

\bottomrule
\end{tabular}
\caption{Student training details and hyperparameters.}
\label{table:student-training}
\end{table}
\end{center}

\end{document}